\documentclass{article}
\usepackage{fancyhdr}
\usepackage{ijcai21}

\usepackage{times}
\usepackage{soul}
\usepackage{url}
\usepackage[hidelinks]{hyperref}
\usepackage[utf8]{inputenc}
\usepackage[small]{caption}
\usepackage{graphicx}
\usepackage{amsmath}
\usepackage{amsthm}
\usepackage{amssymb}
\usepackage{bm}
\usepackage{mathrsfs}
\usepackage{booktabs}
\usepackage{algorithm}
\usepackage{algorithmic}
\usepackage[table]{xcolor}
\usepackage[T1]{fontenc}

\pubnote{\small{Proceedings of the Thirtieth International Joint Conference on Artificial Intelligence (IJCAI-21)}}

\title{Learning Attributed Graph Representations with Communicative Message Passing Transformer}

\author{
Jianwen Chen$^1$\footnotemark[2]
\and
Shuangjia Zheng$^{1,4}$\footnotemark[2]\footnotemark[1]
\and
Ying Song$^2$
\and
Jiahua Rao$^{1,4}$
\And
Yuedong Yang$^{1,3}$\footnotemark[1]
\affiliations
$^1$School of Computer Science and Engineering, Sun Yat-sen University \\
$^2$School of Systems Science and Engineering, Sun Yat-sen University \\
$^3$Key Laboratory of Machine Intelligence and Advanced Computing, Sun Yat-sen University\\
$^4$Galixir Technologies Ltd, Beijing
\emails
\{chenjw48, zhengshj9, songy75, raojh6\}@mail2.sysu.edu.cn, yangyd25@mail.sysu.edu.cn
}

\begin{document}

\maketitle
\renewcommand{\thefootnote}{\fnsymbol{footnote}}
\footnotetext[2]{These two authors contributed equally.}
\footnotetext[1]{Corresponding authors.}
\footnotetext[3]{https://github.com/jcchan23/CoMPT}

\begin{abstract}
  Constructing appropriate representations of molecules lies at the core of numerous tasks such as material science, chemistry and drug designs. Recent researches abstract molecules as attributed graphs and employ graph neural networks (GNN) for molecular representation learning, which have made remarkable achievements in molecular graph modeling. Albeit powerful, current models either are based on local aggregation operations and thus miss higher-order graph properties or focus on only node information without fully using the edge information. For this sake, we propose a Communicative Message Passing Transformer (CoMPT) neural network to improve the molecular graph representation by reinforcing message interactions between nodes and edges based on the Transformer architecture. Unlike the previous transformer-style GNNs that treat molecules as fully connected graphs, we introduce a message diffusion mechanism to leverage the graph connectivity inductive bias and reduce the message enrichment explosion. Extensive experiments demonstrated that the proposed model obtained superior performances (around 4$\%$ on average) against state-of-the-art baselines on seven chemical property datasets (graph-level tasks) and two chemical shift datasets (node-level tasks). Further visualization studies also indicated a better representation capacity achieved by our model.
\end{abstract}

\section{Introduction}

Accurate characterization of molecular properties remains largely an open challenge, and its solution may unlock a widespread use of deep learning in the drug discovery industry \cite{wu2018moleculenet}. Traditionally, this has involved translating a molecule $m$ to a dense feature vector with a representation function, $h = g(m)$, and then applying a variety of techniques to predict the targeted property based on the representation by $y = f(h)$.

Early predictive modeling methods such as quantitative structure-property relationships (QSPR) have been performed based on fixed representations such as expert-crafted physicochemical descriptors and molecular fingerprints \cite{rogers2010extended}. However, descriptor-based methods presume that all target property-related information is covered by the chosen descriptor set, limiting the capability for a model to make problem-specific decisions.

More naturally, a molecular structure can be abstracted as a topological graph with attributed nodes and edges, where node features correspond to atom properties like atomic identity and degree, edge features correspond to bond properties, like bond type and aromaticity. In this sense, graph representation models, especially Graph Neural Networks (GNN), can be intuitively introduced to learn the representations of molecules. Generally, the procedure of GNN framework can be summarized in three main steps: (1) Initialization step, where nodes are initialized with their initial attributes or structural features; (2) Message Passing step, where the features at each node are transmitted from its neighbors across the molecular graph into a message vector; (3) Read-out step, where the node messages are aggregated or pooled into a fixed-length feature vector. Under the above framework, many GNN architectures have been proposed for effective graph representation learning, that achieve promising results in many property prediction tasks \cite{duvenaud2015convolutional,yang2019learned,song2020communicative}.

Despite the fruitful progress, several issues still impede the performance of the current GNN in the molecular graph. First, common graph convolutional operations aggregate only local information and suffer from the suspended animation problem when stacking excessive GNN layers \cite{zhang2019gresnet}, so these models are naturally difficult to learn long-range dependencies and the global chemical environment of each atom. Second, main-stream GNN and its variants mainly focus on obtaining effective nodes embedding but weaken the information carried by edges that is also important for informative graph representations \cite{shang2018edge}. Meanwhile, the node representations obtained by such deep models tend to be over-smoothed and hard to distinguish \cite{li2018deeper}. Such issues greatly hinder the applications of GNNs for molecular representation learning tasks.

To address the above problems, many efforts have been made from different directions. On the one hand, with the emerging of Transformer \cite{vaswani2017attention} in sequence modeling, several Transformer-style GNNs \cite{chen2019path,maziarka2020molecule} have been introduced to learn the long-range dependencies in graph-structured data. These methods can be viewed as a variant of the Graph Attention Network (GAT) \cite{velivckovic2017graph} on a fully connected graph constructed by all atoms, which ignore the graph connectivity inductive bias. As a result, they perform poorly in the tasks where graph topology plays an important rule. On the other hand, the directed message passing neural network \cite{yang2019learned} and its variants \cite{song2020communicative} have been proposed to transmit messages through directed edges rather than vertices. Such methods make use of the edge information explicitly and avoid unnecessary loops in the message passing trajectory, but they still cannot deal with the long-range dependencies.

Based on these observations, we propose a Communicative Message Passing Transformer (CoMPT) neural network for molecular representation learning. In contrast to the previous Transformer-style GNNs that emphasize the node information, CoMPT invokes a communicative message-passing paradigm by strengthening the message interactions between edges and nodes. In our framework, both the edge and node embeddings are updated during the training process. Besides, we refine the message passing process by using the topological connection matrix with a diffusion mechanism to reduce the message enrichment explosion. By selectively propagating information within a molecular graph, CoMPT is able to extract more expressive representation for down-stream tasks.
The main contributions of this work include:
\begin{itemize}
  \item We propose a novel communicative message passing transformer, namely CoMPT, that explicitly captures the atom and bond information of molecular graphes and incorporates both local and global structural information.
  \item Our model includes an elegant way to fuse topology connection matrix using the message diffusion mechanism inspired by the thermal diffusion phenomenon, which was further demonstrated to alleviate the over-smoothing problem. 
  \item Numerical experiments are conducted on both graph-level and node-level public datasets to demonstrate the effectiveness of our method. CoMPT surpasses the state-of-the art models on all nine tasks by up-to 4\% improvement in the average performance.
\end{itemize}

\section{Related Work}
\paragraph{Molecular representation learning.} One of the most popular representations of molecules is the fixed representations through chemical fingerprints, such as Extended Connectivity Fingerprints (ECFP) \cite{rogers2010extended} and chemical descriptors. The heuristics integrated in descriptor generation algorithms typically embed high-level chemistry principles, attempting to maximize the information content of the resulting feature vectors. While these methods can be clearly successful, they always feature a trade-off by emphasizing certain molecular features, while neglecting others. The selections of features are hard-coded in the algorithm and not amenable to problem-specific tuning. Recent works started to explore the molecular graph representation. Early studies learned to only encode the node features \cite{duvenaud2015convolutional} without considering bond information. To gain supplementary information from edges, \cite{kearnes2016molecular} proposed to utilize attributes of both atoms and bonds, and \cite{gilmer2017neural} summarized it into a MPNN framework. Though a few more studies used the information of the edges through network modules such as the edge memory module \cite{withnall2020building}, these models were mainly built upon the node-based MPNN and thus still suffered from the information redundancy during message aggregations. DMPNN \cite{yang2019learned} was introduced as an alternative as it abstracted the molecular graph as an edge-oriented directed graph, avoiding the unnecessary loops in message passing procedure. CMPNN \cite{song2020communicative} further extend this work by strengthening the message interactions between nodes and edges through a communicative kernel. Our work is closely related to CMPNN, while our model built on Transformer is more elegant to capture long-range dependencies and structural variety.

\paragraph{Transformer-style graph neural network.} Several attempts have been made to integrate transformer and graph neural network. \cite{chen2019path} introduced the Path-Augmented Graph Transformer Networks to explicitly take account for longer-range dependencies in molecular graph. One closely related work is \cite{maziarka2020molecule}, which proposed a Molecule Attention Transformer by augmenting the attention mechanism in Transformer using inter-atomic distances and the molecular graph structure. Another work worth to note is GROVER \cite{rong2020self}, which provided a self-supervised pre-trained transformer-style message passing neural network for molecular representation learning. While our model also builds on Transformer \cite{vaswani2017attention} to encode graphs, we contribute new techniques to leverage the graph connectivity inductive bias.

\section{Methods}

In this section, we first briefly review basic concepts of the Transformer model \cite{vaswani2017attention}. Then, we focus on our contributions, describing our alternative in the Transformer encoder framework that uses node-edge message interaction module instead of the self-attention mechanism to pass the message and learn expressive representations for attributed molecular graphs. Finally, we introduce a message diffusion mechanism for leveraging the graph connectivity inductive bias and reducing the message enrichment explosion.

\begin{figure}[!h]
\centering
\includegraphics[scale=0.25]{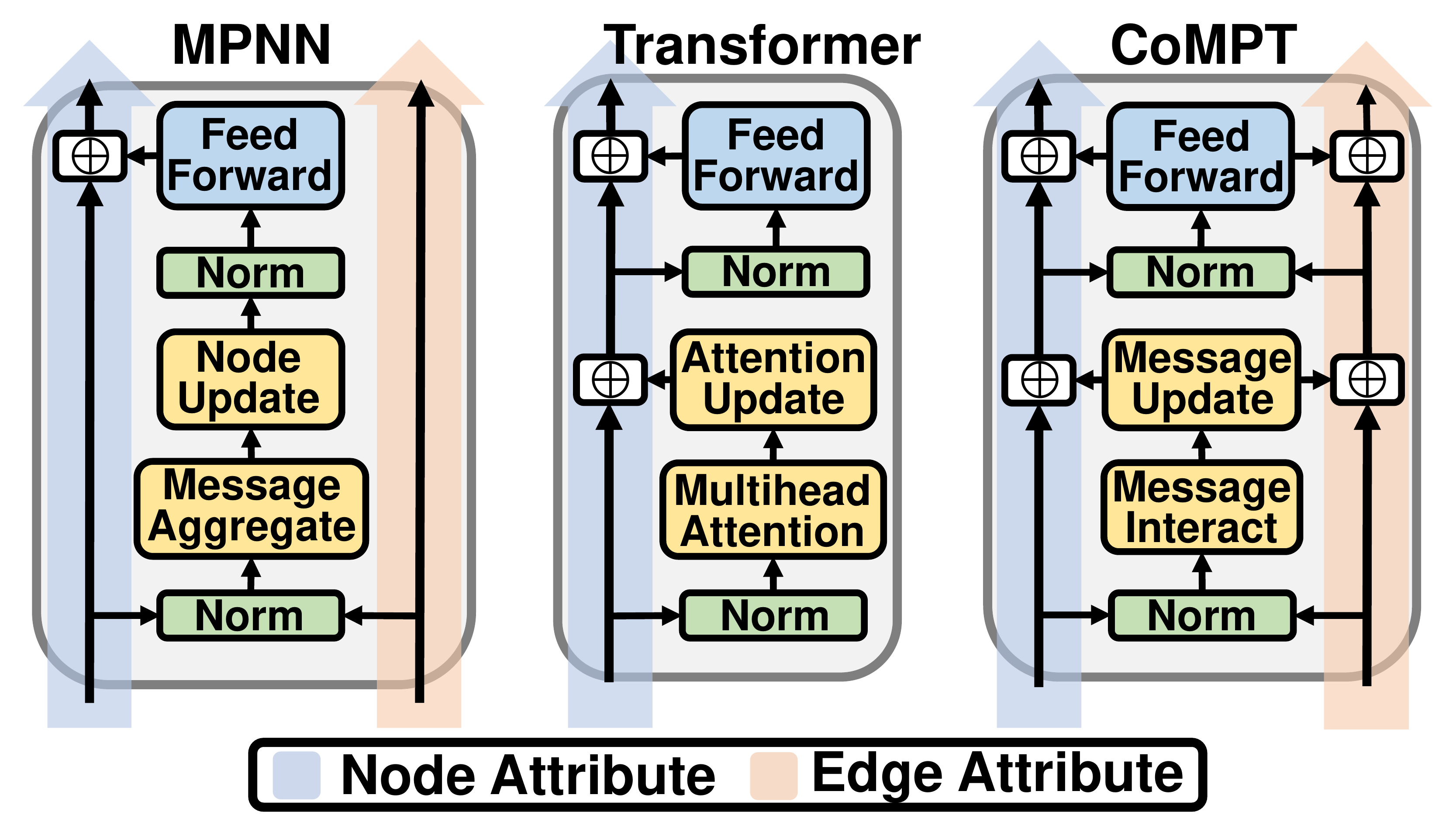}
\caption{Comparing  message  passing  procedure among MPNN (left), Transformer (middle) and CoMPT (right).}
\label{fig:Model}
\end{figure}

\subsection{Preliminary}

\paragraph{Notation and problem definition.} A molecular structure can be considered as an attributed graph $ G=(\mathcal{V},\mathcal{E}) $, where $ |\mathcal{V}|=n$ denotes a set of $n$ atoms (nodes) and $ |\mathcal{E}|=m $ denotes a set of $m$ bonds (edges). $\mathcal{N}_{v}$ is utilized to denote the set of node $v$'s neighbors. For each node and edge, we use $f_{node}$ and $f_{edge}$ represents the feature dimensions, respectively. Following \cite{yang2019learned} that passing message with directed edges, we treat molecular structures as directed graphs to avoid messages being passed along any unnecessary loops in the aggregation procedure. As such, we use $ x_{v} \in \mathbb{R}^{f_{node}} $ to represent the initial features of node $v$, and $e_{uv} \in \mathbb{R}^{f_{edge}}$ are the initial features of the edge $(u,v)$ with direction $u \rightarrow v$. $X \in \mathbb{R}^{n \times f_{node}}$ and $E \in \mathbb{R}^{n \times n \times f_{edge}}$ are the matrices form for all nodes and edges. Besides, there are generally two categories of supervised tasks in the molecular graph learning problems: i) \emph{Graph classification/regression}, where a set of molecular graphs $\{ G_{1},…,G_{N} \}$ and their labels/targets $\{ y_{1},…,y_{N} \}$ are given, and the task is to predict the label/target of a new graph. ii) \emph{Node classification/regression}, where each node $v$ in multiple graphs has a label/target $y_{v}$, and the task is to predict the labels/targets of nodes in the unseen graphs.

\paragraph{Attention mechanism.} Our CoMPT model are built on the transformer encoder framework, in which the attention module is the main building block. The usual implementation of the attention module is the dot product self-attention, which takes inputs with a set of queries, keys, and values ${(q,k,v)}$ that are projected from hidden node features $h(X)$. Then it computes the dot product of the query with all keys and applies a \emph{softmax} function to obtain weights on the values. By stacking the set of ${(q,k,v)}$ s into matrices ${(Q,K,V)}$, it allows highly optimized matrix multiplication operations. Specifically, the outputs can be formulated as:
\begin{equation}
\left \{
\begin{array}{l}
[Q, K, V] = h(X)[W^{Q}, W^{K}, W^{V}] \\
Attention(Q,K,V) = softmax(QK^{T} / \sqrt{d})V
\end{array}
\right .
\end{equation}
where $d$ is the dimension of ${q}$ and ${k}$. Furthermore, we focus on the multi-head attention, where $l$ attention layers are stacked together. The output matrix can be extended as,
\begin{small}
\begin{equation}
\left \{
\begin{array}{l}
Multihead(Q,K,V) = Cat(head_{1},…,head_{l})W^{O} \\
head_{i} = Attention(h(X)W^{Q}_{i}, h(X)W^{K}_{i}, h(X)W^{V}_{i})
\end{array}
\right .
\end{equation}
\end{small}
where ${W^{Q}_{i}, W^{K}_{i}, W^{V}_{i}}$ are the projection matrices of head $i$.

\subsection{The Framework of CoMPT}

\paragraph{Encoding for node position and edge direction.} Compared to the Transformer encoder that only takes node attributes, our CoMPT takes three inputs: node features $X$, edge features $E$, and the topology connection matrix $A \in \mathbb{R}^{n \times n}$ that is computed by using the length of the shortest path between any two nodes. Since the initial node and edge features (more details are listed in the Appendix) do not involve information related to the node position and the edge direction, we need to add annotations explicitly with these features before being fed into the CoMPT model. Specifically, for any node $v_{i} (i = 1,2,…,n)$ and its corresponding initial feature $x_{i}$, we train a learnable position embedding vector $pos_{i}$ according to the atomic index of the node, and then add the initial features to get the hidden features, which can be formulated as:
\begin{equation}
h(x_{i})=node \_ embedding(x_{i}) + pos_{i}
\end{equation}
where $node \_ embedding()$ projects the initial feature to the corresponding dimension. For any directed edge $e_{uv}$, we followed the previous method in \cite{song2020communicative} by adding the source node feature to the initial feature. It could be formulated as:
\begin{equation}
h(e_{uv})=edge \_ embedding(e_{uv}) + h(x_{u})
\end{equation}
where $edge \_ embedding()$ also projects the initial feature to the corresponding dimension. For convenience, we set the same dimension $f$ for all hidden features.

\paragraph{Node-Edge message interaction module.} The key idea behind CoMPT is that we use hidden node features $h(X)$ and hidden edge features $h(E)$ to compute a message interaction scores $M$, which replaces the self-attention scores in the original encoder layer. Specifically, three matrices $Q,K,V$ are firstly calculated by the formula:
\begin{equation}
\left \{ 
\begin{array}{l} 
[Q,V]=h(X)[W^{Q}, W^{V}] \\
K=h(E)W^{K} \\
\end{array}
\right .
\end{equation}
where $W^{Q}, W^{K}, W^{V}$ are the projection matrices. The message interaction matrix $T$ is generated by the inner product:
\begin{equation}
T=matmul(Q,K.tranpose(-2,-1))
\end{equation}
or equivalently in each position
\begin{equation}
T[i,u,v]=matmul(q_{i}, k_{uv})
\end{equation}
where $q_{i}$ and $k_{uv}$ denote the hidden vector of node $i$ and directed edge $(u,v)$, respectively. The intuition behind this tensor product is straight-forward: we compute the scalar product between each node and edge in order to generate the structural message related to the molecule for the final prediction. Subsequently, a selection step is applied to the message interaction matrix $T$ to preserve the molecular graph connectivity. Here, we select three types of matrices according to the node’s neighbors: the node interacts with its outgoing edges ($\bm{M_{o}}$), incoming edges ($\bm{M_{i}}$) and the self-loop edge ($\bm{M_{d}}$). For convenience, the generation and the selection step could be merged by \emph{einsum} operation with the formula below:
\begin{equation}
\left \{
\begin{array}{l}
M_{o}=einsum_{nf,nmf \rightarrow nm}(Q,K) \\
M_{i}=einsum_{nf,mnf \rightarrow nm}(Q,K) \\
M_{d}=diag(M_{o}) = diag(M_{i})
\end{array}
\right .
\end{equation}
After computing three matrices, we normalize them to sum up to 1.0 in each row with the \emph{softmax} function $\sigma$, and compute the final message by:
\begin{equation}
M=\sigma (M_{o}) + \sigma (M_{i}) - \sigma (M_{d})
\end{equation}
This operation eliminates the double-counted self-loop message and avoids the information explosion.
This final message is further utilized to update the node hidden features $h(X)$ and edge hidden features $h(E)$ in each encoder layer. Furthermore, in the situation of multi-head attention, similar to the original transformer framework, we also stack all $l$ attention blocks at the end of each layer.

\paragraph{Residual message update module.} In the original transformer encoder framework, the self-attention scores are utilized to compute the weighted sum of vectors for each node, which could be regarded as an updated operation that is applied to all nodes. In contrast, we utilize the message interaction scores $M$ to update the node hidden features $h(X)$. Besides, inspired by \cite{song2020communicative}, the edge hidden features are also updated according to the rich information that comes from $M$. The update operation is concluded as:
\begin{equation}
\left \{
\begin{array}{l}
h(X)=matmul(M, V) \\
h(E)=M\odot K
\end{array}
\right .
\end{equation}
where $\odot$ denotes the element-wise operation. Besides, CoMPT model has multiple stacked layers, where each encoder layer consists of a multi-head message interaction module, message update module and a position-wise forward module. To make the training step more stable, we adopt the post layer norm module \cite{xiong2020layer} before getting into each module. Furthermore, residual connections between any two encoder layers are added for reducing the vanishing of the gradient, which can be formulated as:
\begin{equation}
\left \{
\begin{array}{l}
h_{k+1}(X)=h_{k}(X)+Encoder(h_{k}(X), h_{k}(E)) \\
h_{k+1}(E)=h_{k}(E)+Encoder(h_{k}(X), h_{k}(E))
\end{array}
\right .
\end{equation}
where $k$ represents the index of the encoder layer, $Encoder(\cdot)$ denotes the whole encoder layer with the various modules mentioned above.

\subsection{Message Diffusion and Global Pooling}
The key to accurately predict the properties on the graph-level/node-level tasks is how to keep the message interacting correctly. Previous studies have shown that deep message aggregation of GNN will lead to an over-smooth phenomenon\cite{li2018deeper}, where the features of nodes within the graph will converge to the same values. 

To alleviate this issue, we design a simple attenuation mechanism for message passing during iteration to delay the process of aggregating redundant information to nodes. In particular, each hidden vector in message interaction scores $M$ could be regarded as the prepared message sent from the row index of the node to the column index of the node, and the topology connection matrix $A$ shows the distance of shortest path between two nodes. We add the attenuation coefficient by using the Gaussian kernel function with the formula:
\begin{equation}
M(u, v) = M(u, v)e^{-\alpha A(u,v)}
\end{equation}
where $M(u, v)$ denotes the message sending from $u$ to $v$, $A(u,v)$ means the shortest path between $u$ and $v$, $\alpha \in [0,1]$ is a trainable coefficient to control the attenuation level. It is obvious that with the increase of distance, the message will decay rapidly in the beginning, and then turn smooth for a long distance. After applying this mechanism, we can defer the over-smoothness of the aggregating process to a certain degree. The ablation study in the section 4.3 also shows that the attenuation mechanism improves the prediction performance.

Finally, for the graph-level tasks, a readout operator/generation layer is added to obtain a fixed feature vector for the molecule. Here, we adopt a Gated Recurrent Unit (GRU) for global pooing following \cite{gilmer2017neural,song2020communicative} as:
\begin{equation}
z=\sum_{x \in \mathcal{V}}GRU(h(x))
\end{equation}
where $h(x)$ is the set of atom representations in the molecular graph, and $GRU$ is the Gated Recurrent Unit. Finally, we perform downstream property prediction $\hat{y} = f(h)$ where $f(\cdot)$ is a fully connected layer.

\begin{table*}
\small
\centering
\setlength{\tabcolsep}{1mm}{
\begin{tabular}{c|c|c|c|c|c|c|c}
\hline
Task & 
\multicolumn{4}{c|}{Graph Classification(ROC-AUC)} & 
\multicolumn{3}{c}{Graph Regression(RMSE)}\\
\hline
Dataset  & BBBP & Tox21 & Sider & ClinTox
& ESOL & FreeSolv & Lipophilicity \\
\hline
TF\_Robust
& 0.860 $\pm$ 0.087 & 0.698 $\pm$ 0.012 & 0.607 $\pm$ 0.033 & 0.765 $\pm$ 0.085
& 1.722 $\pm$ 0.038 & 4.122 $\pm$ 0.085 & 0.909 $\pm$ 0.060 \\
GCN 
& 0.877 $\pm$ 0.036 & 0.772 $\pm$ 0.041 & 0.593 $\pm$ 0.035 & 0.845 $\pm$ 0.051 
& 1.068 $\pm$ 0.050 & 2.900 $\pm$ 0.135 & 0.712 $\pm$ 0.049 \\
Weave 
& 0.837 $\pm$ 0.065 & 0.741 $\pm$ 0.044 & 0.543 $\pm$ 0.034 & 0.823 $\pm$ 0.023 
& 1.158 $\pm$ 0.055 & 2.398 $\pm$ 0.250 & 0.813 $\pm$ 0.042\\
SchNet 
& 0.847 $\pm$ 0.024 & 0.767 $\pm$ 0.025 & 0.545 $\pm$ 0.038 & 0.717 $\pm$ 0.042 
& 1.045 $\pm$ 0.064 & 3.215 $\pm$ 0.755 & 0.909 $\pm$ 0.098\\
N-Gram
& 0.912 $\pm$ 0.013 & 0.769 $\pm$ 0.027 & \cellcolor{gray!40} 0.632 $\pm$ 0.005 & 0.855 $\pm$ 0.037 
& 1.100 $\pm$ 0.160 & 2.512 $\pm$ 0.190 & 0.876 $\pm$ 0.033 \\
AttentiveFP 
& 0.908 $\pm$ 0.050 & 0.807 $\pm$ 0.020 & 0.605 $\pm$ 0.060 & \cellcolor{gray!40} 0.933 $\pm$ 0.020 
& 0.853 $\pm$ 0.060 & 2.030 $\pm$ 0.420 & 0.650 $\pm$ 0.030 \\
\hline
MPNN 
& 0.913 $\pm$ 0.041 & 0.808 $\pm$ 0.024 & 0.595 $\pm$ 0.030 & 0.879 $\pm$ 0.054 
& 1.167 $\pm$ 0.430 & 2.185 $\pm$ 0.952 & 0.672 $\pm$ 0.051 \\
MGCN
& 0.850 $\pm$ 0.064 & 0.707 $\pm$ 0.016 & 0.552 $\pm$ 0.018 & 0.634 $\pm$ 0.042 
& 1.266 $\pm$ 0.147 & 3.349 $\pm$ 0.097 & 0.650 $\pm$ 0.030\\
DMPNN 
& 0.919 $\pm$ 0.030 & \cellcolor{gray!40} 0.826 $\pm$ 0.023 & \cellcolor{gray!40} 0.632 $\pm$ 0.023 & 0.897 $\pm$ 0.040 
& 0.980 $\pm$ 0.258 & 2.177 $\pm$ 0.914 & 0.653 $\pm$ 0.046 \\
CMPNN 
& \cellcolor{gray!40} 0.927 $\pm$ 0.017 & 0.806 $\pm$ 0.016 & 0.616 $\pm$ 0.003 & 0.902 $\pm$ 0.008 
& \cellcolor{gray!40} 0.798 $\pm$ 0.112 & 2.007 $\pm$ 0.442 & \cellcolor{gray!40} 0.614 $\pm$ 0.029 \\
\hline
Smiles Transformer
& 0.900 $\pm$ 0.053 & 0.706 $\pm$ 0.021 & 0.559 $\pm$ 0.017 & 0.905 $\pm$ 0.064 
& 1.144 $\pm$ 0.118 & 2.246 $\pm$ 0.237 & 1.169 $\pm$ 0.031 \\
GROVER 
& 0.911 $\pm$ 0.008 & 0.803 $\pm$ 0.020 & 0.624 $\pm$ 0.006 & 0.884 $\pm$ 0.013 
& 0.911 $\pm$ 0.116 & \cellcolor{gray!40} 1.987 $\pm$ 0.072 & 0.643 $\pm$ 0.030 \\
\hline
CoMPT 
& {\bf 0.938 $\pm$ 0.021} & \underline{0.809 $\pm$ 0.014} & {\bf 0.634 $\pm$ 0.030} & {\bf 0.934 $\pm$ 0.019}
& {\bf 0.774 $\pm$ 0.058} & {\bf 1.855 $\pm$ 0.578} & {\bf 0.592 $\pm$ 0.048} \\
\hline
\end{tabular}}
\caption{Prediction results of CoMPT and baselines on seven chemical graph datasets. We used a 5-fold cross validation with scaffold split and replicated experiments on each tasks for five times. Mean and standard deviation of AUC or RMSE values are reported.}
\label{tab:Graph level scaffold Result}
\end{table*}

\section{Experiments}

In this section, we evaluate the proposed model CoMPT on three kinds of tasks. We aim to answer the following research questions:
\begin{itemize}
\item \textbf{RQ1}: How does CoMPT model perform compared with state-of-the-art molecular property prediction methods?
\item \textbf{RQ2}: How do different components (i.e, node position, edge direction, and message diffusion mechanism) affect CoMPT?
\item \textbf{RQ3}: Can CoMPT model provide better representations for the attributed molecular graphs?
\end{itemize}

\subsection{Experiment Setups}

\subsubsection{Benchmark Datasets}
To enable head-to-head comparisons of CoMPT to existing molecular representation methods, we evaluated our proposed model on nine benchmark datasets across three kinds of tasks from \cite{wu2018moleculenet} and \cite{jonas2019rapid}, each kind of which consists of 2 to 4 public benchmark datasets, including BBBP, Tox21, Sider, and ClinTox for Graph Classification tasks, ESOL, FreeSolv and Lipophilicity for Graph Regression tasks, chemical shift prediction of hydrogen and carbon for Node Regression tasks. The statistics of datasets are shown in Table S1.

In the graph-level task, following the previous works, we utilized a 5-fold cross-validation and replicate experiments on each task five times. Note that we adopted the scaffold split method recommended by \cite{yang2019learned} to split the datasets into training, validation, and test, with a 0.8/0.1/0.1 ratio. Scaffold Split is a more challenging and realistic evaluation setting in molecular property prediction tasks by guaranteeing the high molecular scaffold diversity of the training, validation, and test sets. 

In the node-level task, we follow the previous study \cite{jonas2019rapid} by randomly splitting the dataset into 80\% as the training set and 20\% as the test set, and then use 95\% of training data to train the model and the remaining 5\% to validate the model for early stopping. All methods report the mean and standard deviation of corresponding metrics. To improve model performance, we applied the grid search to obtain the best hyper-parameters of the models. 

\subsubsection{Baselines Comparison} 
We comprehensively compared CoMPT against with 12 baseline methods in the graph level task. These models were most shown in the MoleculeNet \cite{wu2018moleculenet} and GROVER as follows: TF\_Robust \cite{ramsundar2015massively} is a DNN-based multitask framework taking the molecular fingerprints as the input. GCN, Weave, and SchNet \cite{duvenaud2015convolutional,kearnes2016molecular} are three graph convolutional models. N-Gram \cite{liu2019n} is a state-of-the-art unsupervised representation method for molecular property prediction. AttentiveFP \cite{xiong2019pushing} is an extension of the graph attention network. MPNN and its variants MGCN \cite{lu2019molecular}, DMPNN and CMPNN are models considering the edge features during message passing. Specifically, to demonstrate the power of the message-interaction module, we also compare CoMPT with two transformer model: Smiles Transformer \cite{honda2019smiles} and GROVER. For a fair comparison, we only report the results without the pretrained strategy.

In the node level task, we compared our CoMPT model with the other 3 proposed methods in this benchmark. The first one is HOSE codes, which attempted to summarize the neighborhood around each atom in concentric spaces, and then use a nearest-neighbor approach to predict the particular shift value. The rest baselines include GCN \cite{jonas2019rapid} and MPNN \cite{kwon2020neural}, where they used different deep graph neural networks to improve the performance of prediction.

\subsection{Performance Comparison (RQ1)}

\begin{figure*}[!h]
\centering
\includegraphics[scale=0.6]{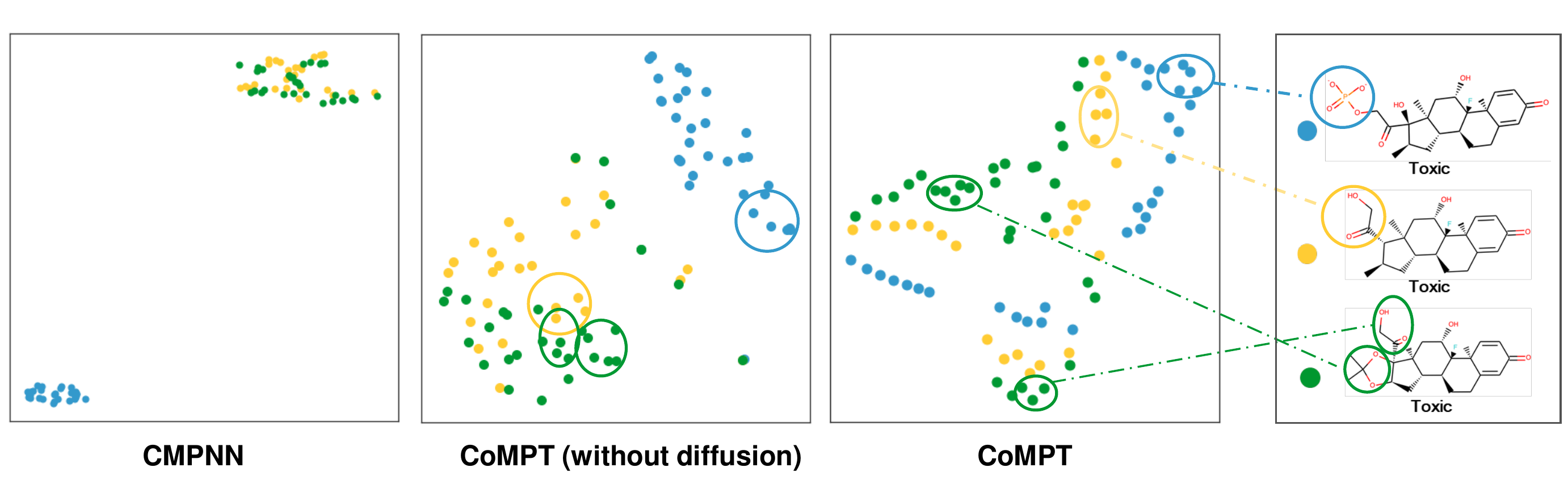}
\caption{T-SNE visualization of atom embeddings for three similar molecules (in three colors) that have a common scaffold but various side-chains in ClinTox dataset. Ideally, the scaffold atom embeddings of these three molecules should be mixed together while the unique side-chains’ embeddings should be distinguishable.}
\label{fig:Graph-level}
\end{figure*}

\paragraph{Performance in graph level task.} Table \ref{tab:Graph level scaffold Result} displays the complete results of each model on all datasets, where cells in the gray shadow denote the previous best methods, and cells with the bold style show the best result achieved by CoMPT. Table \ref{tab:Graph level scaffold Result} presents some observations: (1) Both the message passing neural network and transformer framework perform better than graph neural network on most datasets, and CoMPT combines the advantages of them to achieve the best performances on 6 out of 7 datasets. Compared to the previous best message passing method CMPNN and transformer method GROVER, the general improvements are 3.4\% (2.0\% on classification tasks and 3.4\% on regression tasks) and 4.7\% (2.7\% on classification tasks and 4.7\% on regression tasks), respectively. This notable incresing suggests the effectiveness of the structural representation learned by CoMPT for graph level prediction tasks. (2) The message passing neural network performs better than the transformer neural network, indicating the importance of edge features relative to only an adjacency matrix or distance matrix. (3) In the situation of small dataset, such as the Freesolv task with only 642 labeled molecules, CoMPT gains a 6.6\% relative improvement over previous SOTAs, confirming that CoMPT model could enhance the performance on the task with few labeled data.

\paragraph{Performance in node level task.} Table \ref{tab:Node level Result} shows the comparison results of the baseline and CoMPT on the prediction of 1H-NMR and 13C-NMR spectra in terms of MAE. We performed experiments 5 times independently with different random seeds and report the average and standard deviation over the 5 repetitions. The average values over the 5 repetitions for 1H-NMR and 13C-NMR are 0.214 and 1.321 ppm per NMR-active atom, respectively. This indicates that our CoMPT model could extract the meaningful latent node representations and thus enable more accurate predictions of NMR spectra for new molecules.

\begin{table}
\centering
\setlength{\tabcolsep}{3mm}{
\begin{tabular}{c|c|c}
\hline
Task & \multicolumn{2}{c}{Node Regression(MAE)} \\
\hline
Dataset & 1H-NMR & 13C-NMR \\
\hline
HOSE & 0.33 & 2.85 \\
GCN  & 0.28 & 1.43 \\
MPNN & 0.229 $\pm$ 0.002 & 1.355 $\pm$ 0.022 \\
CoMPT & {\bf0.214 $\pm$ 0.003} & {\bf 1.321 $\pm$ 0.012} \\
\hline
\end{tabular}}
\caption{Performance on Node-level tasks}
\label{tab:Node level Result}
\end{table}

\begin{table}
\small
\centering
\setlength{\tabcolsep}{0.6mm}{
\begin{tabular}{c|c|c|c}
\hline
Dataset & ClinTox & Lipophilicity & 1H-NMR \\
\hline
Without All & 0.862 & 0.653 & 0.231\\
Without message diffusion & 0.868 & 0.651 &  0.221 \\
Without node position & 0.903 & 0.614 &  0.217 \\
Without edge direction & 0.902 & 0.612 & 0.218 \\
\hline
CoMPT & {\bf 0.934} & {\bf 0.592} & {\bf 0.214}\\
\hline
\end{tabular}}
\caption{Ablation results on three kinds of datasets}
\label{tab: Ablation results}
\end{table}

\subsection{Ablation Study (RQ2)}

We conducted ablation studies on three benchmark datasets to investigate factors that influence the performance of the proposed CoMPT framework. 

As shown in Table \ref{tab: Ablation results}, CoMPT with the node position, edge direction, and message diffusion shows the best performance among all architectures. The exclusion of all three modules in the “without All” variant performed the worst. The exclusion of the message diffusion mechanism caused larger decreases in performances than the ones excluding two other modules, showing the importance of reducing the message enrichment explosion. Additionally, the uses of node position and edge direction are both helpful for the final performance.

\subsection{Atomic Representation Visualization (RQ3)}
As shown in \cite{li2018deeper}, the node embeddings obtained by deep GNNs tend to be over-smoothed and become indistinguishable, while shallow GNNs cannot capture atom positions within the broader context of the molecular graph. To investigate whether the CoMPT alleviated these issues as expected, we used t-distributed stochastic neighbor embedding (t-SNE) to visualize the atom embedding distributions of three similar compounds that have a common scaffold (the four-membered ring) but different side-chains. Ideally, the scaffold atom embeddings of these three molecules should be mixed together while the unique side-chains' embeddings should be distinguishable.

Figure \ref{fig:Graph-level} shows the projected atomic embeddings extracted from different models using the t-SNE with default settings. Overall, three methods provide reasonable results. MPNN inherits the over-smoothness issue from the GNN, making the atom embeddings indistinguishable within the graph . In contrast, both CoMPT models (with or without diffusion)  can scatter the atoms well with distinguishable node embeddings. Relative to  CoMPT without diffusion, CoMPT could exactly mix the scaffold atom embeddings and differentiate the side chains.
More interestingly,  CoMPT can distinguish the same functional groups in different chemical environments  (the Hydroxy Ketones in green and in yellow, respectively). These results suggest that CoMPT can not only alleviate the over-smoothness but also capture better representations within the broader context of the molecular graph as expected.

In the node-level tasks, our CoMPT model reaches 0.214 MAE that many molecules with densely packed 1H-NMR spectra can be resolved at these levels of accuracy. As an example, Figure S1 depicts the structure of the 3-Formylbenzoic acid, which has 6 hydrogen atoms, labeled from 11 to 16 with peaks within the 4-14 ppm range. The small difference between the ground truth and the prediction further proves that our model has a good capability in the node-level tasks.

\section{Conclusions}
In this paper, we propose a Communicative Message Passing Transformer (CoMPT) neural network to improve the molecular representation by reinforcing the message interactions between nodes and edges based on the Transformer model. Further, we introduce a message diffusion mechanism to decay the message enrichment explosion as well as over-smoothness during the message passing process. Extensive experiments demonstrate that our CoMPT model obtains superior performance against state-of-the-art baselines on both graph-level tasks and node-level tasks.

\section*{Acknowledgments}
This work has been supported by the National Key R\&D Program of China(2020YFB0204803), National Natural Science Foundation of China(61772566), Guangdong Key Field R\&D Plan(2019B020228001, 2018B010109006), Introducing Innovative and Entrepreneurial Teams(2016ZT06D211), Guangzhou S\&T Research Plan(202007030010).

\bibliographystyle{named}
\bibliography{ijcai21}

\end{document}


\maketitle
\renewcommand{\thetable}{S\arabic{table}}
\renewcommand\thefigure{S\arabic{figure}}
\section{Dataset Description}

\noindent \textbf{BBBP}: The Blood-brain barrier penetration dataset includes binary labels for 2035 compounds on their permeability properties.\noindent \textbf{Tox21}: The Tox21 dataset was created in the Tox21 data challenge, which contained qualitative toxicity measurements for 7821 compounds on 12 different targets, including nuclear receptors and stress response pathways. \noindent  \textbf{Sider}: The Side Effect Resource grouped drug side-effects for 1379 approved drugs into 27 system organ classes. \noindent  \textbf{ClinTox}: The ClinTox dataset includes 1468 drug molecules that are approved through the FDA and compounds that failed during clinical trials due to toxicity. \noindent  \textbf{ESOL}: The ESOL is a small dataset consisting of water solubility data for 1128 compounds.\noindent  \textbf{FreeSolv}: The Free Solvation Database provides 642 experimental and calculated hydration free energy of small molecules in water. \noindent  \textbf{Lipophilicity}: The Lipophilicity dataset provides the experimental result of octanol/water distribution coefficient ($logD$ at pH 7.4) of 4198 compounds. \noindent  \textbf{1H/13C-NMR}: The chemical shift dataset were collected from nmrshiftdb2. We excluded molecules with more than 100 atoms or failing to pass the sanitizing process in the RDKit. 

\section{Featurization Extraction}
The feature extraction contains three parts: 1) Node feature extraction. 2) Bond feature extraction. 3) Topology connection matrix. We use RDKit to extract all features as the input of CoMPT. Table \ref{tab: Atom features} and Table \ref{tab: Bond features} show the atom and bond features we used in CoMPT.

\begin{table}[h]
\small
\centering
\begin{tabular}{c|c|c|c|c}
\hline
Dataset  & Tasks & Type & Molecule & Metric \\
\hline
BBBP & 1 & GC & 2,035 & ROC-AUC \\
Tox21 & 12 & GC & 7,821 & ROC-AUC \\
Sider & 27 & GC & 1,379 & ROC-AUC \\
ClinTox & 2 & GC & 1,468 & ROC-AUC \\
\hline
ESOL & 1 & GR & 1,128 & RMSE\\
FreeSolv & 1 & GR & 642 & RMSE\\
Lipophilicity & 1 & GR & 4,198 & RMSE\\
\hline
1H-NMR & 1 & NR & 12800 & MAE\\
13C-NMR & 1 & NR & 26859 & MAE\\
\hline
\end{tabular}
\caption{Statistics of datasets. GC for Graph Classification, GR for Graph Regression, NR for Node Regression}
\end{table}

\begin{table}[h]
\small
\centering
\setlength{\tabcolsep}{0.5mm}{
\begin{tabular}{m{2cm}|m{0.6cm}|m{4.5cm}}
\hline
Features & Size & Description\\
\hline
Atom type & 101 & type of atom (e.g C,N,O) \\
Hybridization & 6 & sp, sp2, sp3, sp3d, sp3d2 or unknown\\
Number of H & 1 & number of bond hydrogen atoms\\
Degrees & 1 & number of neighbor atoms \\
Formal Charges & 1 & number of formal charge \\
Valences & 1 & number of valences\\
Gasteiger Charges & 1 & value of Marsilli-Gasteiger partial charges\\
Gasteiger HCharges & 1 & value of Marsilli-Gasteiger hydrogen partial charges\\
Aromaticity & 1 & whether this atom is part of an aromatic system \\
In ring & 1 & whether the atom is part of a ring \\
\hline
\end{tabular}}
\caption{Atom features}
\label{tab: Atom features}
\end{table}

\begin{table}[h]
\small
\centering
\setlength{\tabcolsep}{0.4mm}{
\begin{tabular}{m{2cm}|m{0.6cm}|m{4.5cm}}
\hline
Features & Size & Description\\
\hline
Bond type & 4 & single, double, triple, aromatic \\
Stereo & 6 & none, any, E/Z or cis/trans \\
In ring & 1 & whether the bond is part of a ring \\
conjugated & 1 & whether the bond is conjugated \\
\hline
\end{tabular}}
\caption{Bond features}
\label{tab: Bond features}
\end{table}

\begin{figure}[h]
\centering
\includegraphics[scale=0.3]{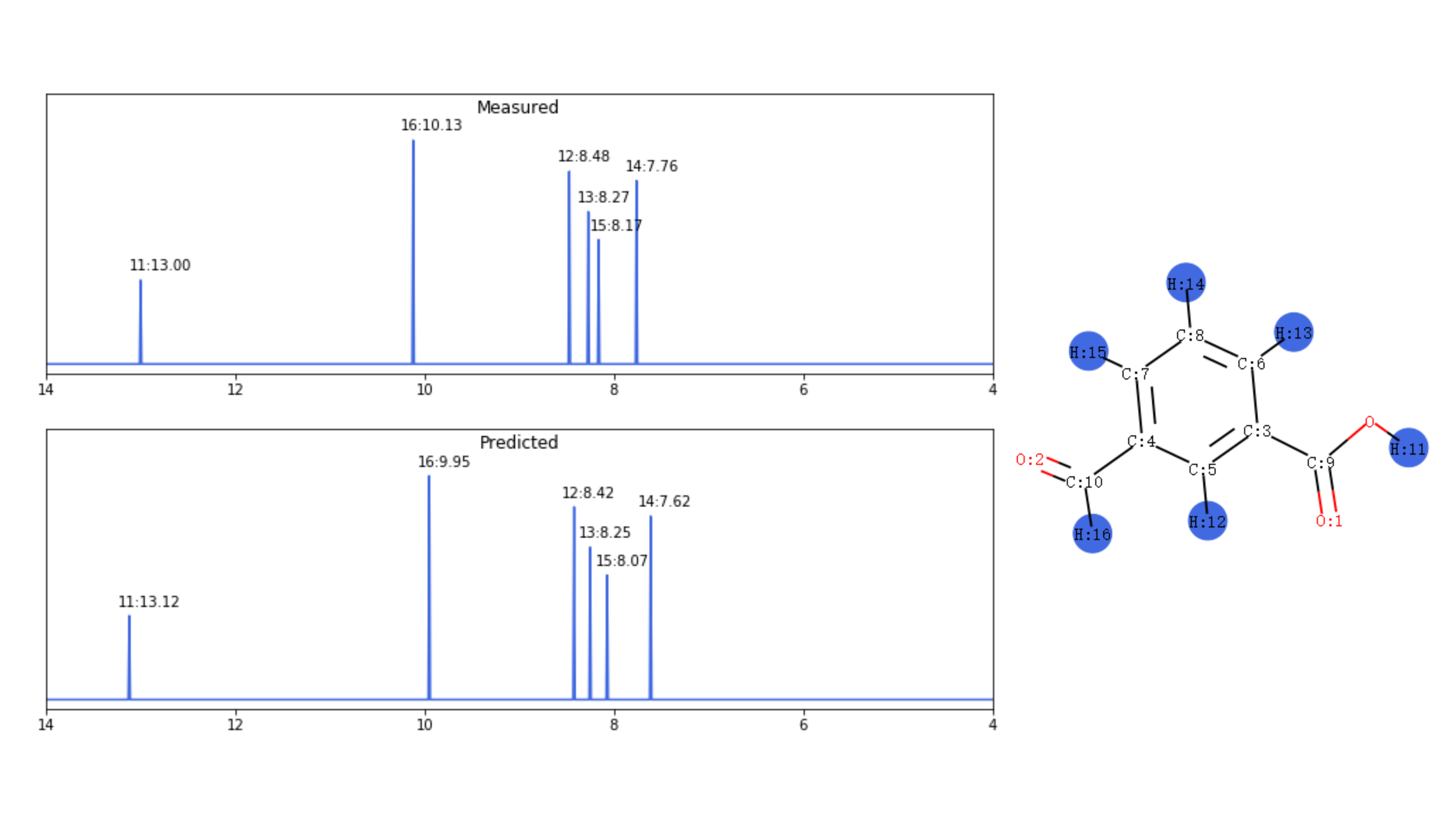}
\caption{Measured and predicted spectra for 3-Formylbenzoic acid}
\label{fig:Graph-level}
\end{figure}
